\documentclass[manuscript,screen]{acmart}

\usepackage{array}

\begin{document}

\title{Culturally-Grounded Governance for Multilingual Language Models: Rights, Data Boundaries, and Accountable AI Design}

\author{Hanjing Shi}
\email{hasa23@lehigh.edu}
\affiliation{
  \institution{Lehigh University}
  \city{Bethlehem}
  \state{Pennsylvania}
  \country{USA}
}

\author{Dominic DiFranzo}
\email{djd219@lehigh.edu}
\affiliation{
  \institution{Lehigh University}
  \city{Bethlehem}
  \state{Pennsylvania}
  \country{USA}
}

\begin{abstract}
Multilingual large language models (MLLMs) are increasingly deployed across cultural, linguistic, and political contexts, yet existing governance frameworks largely assume English-centric data, homogeneous user populations, and abstract notions of fairness. This creates systematic risks for low-resource languages and culturally marginalized communities, where data practices, model behavior, and accountability mechanisms often fail to align with local norms, rights, and expectations.
Drawing on cross-cultural perspectives in human-centered computing and AI governance, this paper synthesizes existing evidence on multilingual model behavior, data asymmetries, and sociotechnical harm, and articulates a culturally grounded governance framework for MLLMs. We identify three interrelated governance challenges: cultural and linguistic inequities in training data and evaluation practices, misalignment between global deployment and locally situated norms, values, and power structures, and limited accountability mechanisms for addressing harms experienced by marginalized language communities.
Rather than proposing new technical benchmarks, we contribute a conceptual agenda that reframes multilingual AI governance as a sociocultural and rights based problem. We outline design and policy implications for data stewardship, transparency, and participatory accountability, and argue that culturally grounded governance is essential for ensuring that multilingual language models do not reproduce existing global inequalities under the guise of scale and neutrality.
\end{abstract}

\ccsdesc[500]{Human-centered computing~Collaborative and social computing}
\ccsdesc[500]{Human-centered computing~Natural language interfaces}
\ccsdesc[300]{Computing methodologies~Natural language processing}
\ccsdesc[300]{Social and professional topics~Cultural characteristics}

\keywords{Multilingual Large Language Models, AI Governance, Trustworthiness, Cross-Cultural Computing, Human-Computer Interaction}

\maketitle

\section{Introduction}

The advent of Large Language Models (LLMs) has profoundly impacted various domains of artificial intelligence, from natural language processing to complex decision-making systems \cite{chang2024survey, tokayev2023ethical}. Initially designed for monolingual tasks, these models have expanded to support multilingual functionalities, giving rise to multilingual LLMs (MLLMs), which cater to the linguistic diversity of a global digital audience \cite{lankford2023adaptmllm, cahyawijaya2024llms}. This transition from traditional machine translation to MLLMs marks a significant advancement, as these models not only translate text between languages but also engage with content across multiple languages simultaneously, offering a comprehensive approach to cross-linguistic understanding \cite{nie2024decomposed, qin2024multilingual}. 

Unlike traditional machine translation systems, which are primarily focused on converting text from one language to another, MLLMs are designed to facilitate collaboration and interaction between languages and cultures. This makes them particularly relevant for fields like Computer-Supported Cooperative Work (CSCW) and Computer-mediated communication (CMC), where interactions often involve participants from diverse linguistic backgrounds. By supporting not only translation, but also nuanced engagement with diverse content, MLLMs have the potential to improve communication, inclusion, and problem solving in multicultural environments \cite{bender2021dangers, adebara2022serengeti}. This opens new avenues for CSCW research, particularly in understanding how MLLMs can influence cooperation, social relationships, and trust across cultural and linguistic barriers.

Despite these capabilities, MLLMs face substantial challenges, especially when it comes to low-resource languages. Of the approximately 7,000 spoken languages and 300 writing systems worldwide, only a small fraction are well represented in the datasets used to train these models \cite{qin2024multilingual}. Consequently, MLLMs often disproportionately benefit high-resource languages such as English, Mandarin, and Spanish, while speakers of low-resource languages experience disparities in model effectiveness \cite{chen2023monolingual, zhang2023don, choudhury2023ask}. Furthermore, pre-training data can embed stereotypes and biases that lead to discriminatory results, perpetuating harmful narratives related to race, gender, and culture \cite{weidinger2022taxonomy}. These biases not only limit the real-world applicability of MLLM, but also raise significant ethical concerns about fairness, inclusion, and cultural sensitivity in diverse settings \cite{tokayev2023ethical, chang2024survey, choudhury2021linguistically}.

This literature review critically examines MLLMs, highlighting that current methodologies can unintentionally contribute to educational, social, and economic inequalities. Specifically, issues such as inadequate tokenization and biased training data often result in misinterpretations and oversimplifications of languages that are underrepresented \cite{ali2023tokenizer, bommasani2021opportunities}. This can have a tangible impact on collaboration, as language users can face barriers in understanding and engaging content that does not capture their cultural nuances. Furthermore, traditional evaluation metrics, such as BLEU scores, are often insufficient for assessing the broader capabilities of MLLMs as collaborative agents, as they focus narrowly on translation accuracy rather than the model's ability to facilitate intercultural understanding.

By positioning MLLMs as collaborative partners rather than mere translators, this review explores innovative applications where seamless cross-linguistic and cross-cultural collaboration is essential. MLLMs are uniquely equipped to adapt to diverse social and cultural contexts, facilitating nuanced communication and enhancing global collaboration through knowledge sharing. Unlike traditional machine translation systems that focus solely on converting text between languages, MLLMs possess the ability to interpret, generate, and engage with content in ways that reflect a deeper cultural and contextual understanding.

To comprehensively assess the trustworthiness of MLLMs, this review focuses on four critical attributes—\textbf{Data Quality}, \textbf{Dependability and Operability}, \textbf{Human-Centered Design}, and \textbf{Human Oversight}. These attributes were chosen due to their importance in supporting the effective deployment of MLLMs within complex, multicultural, and multilingual collaboration settings. The order of these attributes reflects a logical progression in evaluating trustworthiness, starting with foundational technical requirements (Data Quality), moving through operational reliability (Dependability and Operability), addressing user-focused considerations (Human-Centered Design), and culminating in ethical and regulatory oversight (Human Oversight). This structure aligns with the metamodel proposed by Mattioli et al. \cite{mattioli2024overview}, which provides a structured framework for assessing trustworthiness in machine learning systems. Further explanation of each attribute’s role and its implications for MLLMs is detailed in Section 3.2.

Mattioli et al.'s meta-model identifies key dimensions of trustworthiness that apply not only to technical performance but also to the social and ethical aspects of machine learning systems. By evaluating MLLMs through these four lenses, this review not only provides a framework for understanding their current capabilities, but also identifies areas where further research and development are necessary. These attributes are particularly relevant to understanding how MLLMs can foster effective cross-linguistic collaboration while also ensuring that their deployment in multilingual and multicultural settings is equitable, ethical and culturally aware.

The findings of this literature review highlight several critical areas for improvement in the development and deployment of MLLMs:
\begin{enumerate}
    \item The need for culturally sensitive AI systems that can navigate the complexities of cross-cultural communication and collaboration, ensuring inclusivity for speakers of high- and low-resource languages.
    \item Lack of robust evaluation metrics that go beyond technical performance to assess the social and cultural impacts of MLLMs in collaborative real-world settings.
    \item The necessity for tools and methodologies that facilitate trustworthy, AI-mediated communication in diverse group interactions, ensuring that AI systems are not only functional but also ethically sound and reliable in fostering collaboration.
    \item Gaps between current MLLM research and practical applications, particularly in settings where collaboration spans multiple languages and cultural contexts, indicate the need for more interdisciplinary research and real-world testing.
\end{enumerate}

Unlike existing MLLM surveys, such as those by Qin et al.\cite{qin2024multilingual} and Xu et al. \cite{xu2024survey}, which primarily focus on technical innovations such as multilingual alignment strategies, resource availability, and addressing bias within datasets, this review takes a different approach by examining MLLMs through the lens of \textit{collaboration and communication}. Our emphasis on \textit{multicultural, not just multilingual} collaboration fills a critical gap in the current body of work, underscoring the importance of MLLMs in fostering effective cross-linguistic teamwork in practical applications, particularly within the CSCW and HCI domains. By bridging these research gaps, our review aims to offer practical insights that advance the design and implementation of MLLM to support reliable, inclusive and reliable collaboration in a multicultural world.

Figure 1 below provides a visual representation of the review process and the conceptual framework, illustrating how the four key attributes interact with the challenges of MLLM, including biases in low-resource languages, inconsistencies in translation, inability to capture cultural nuances, and trustworthiness in real-world scenarios.

\begin{figure}[ht]
  \centering
  \includegraphics[width=\linewidth]{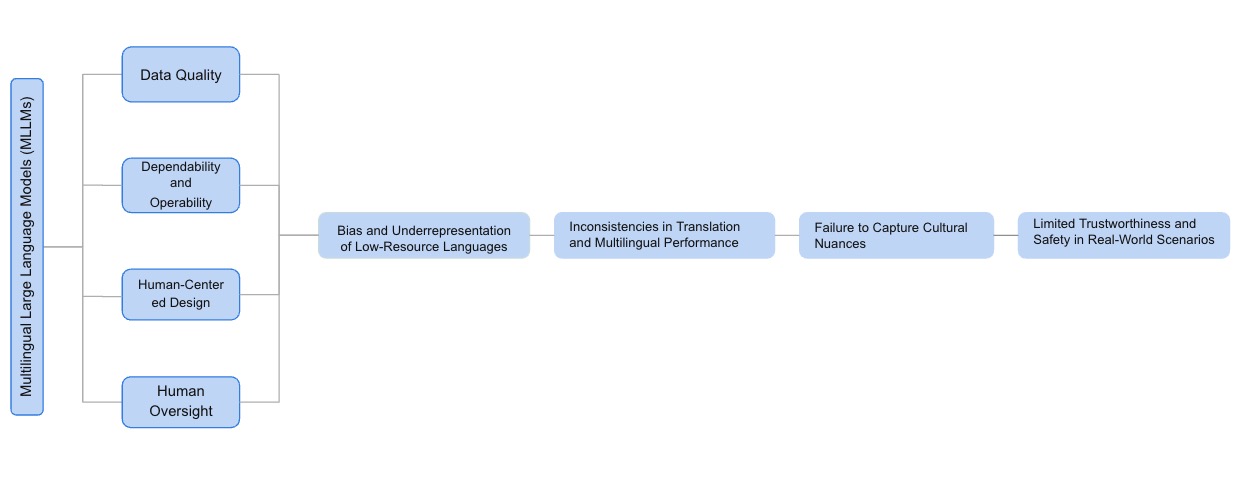}
  \caption{A hierarchical flow illustrating how four key attributes of MLLMs contribute to challenges like bias, translation inconsistencies, and cultural nuances, ultimately affecting trustworthiness and safety in real-world applications.}
  \Description{A flow diagram showing the relationship between key attributes of MLLMs and the resulting challenges, leading to trustworthiness and safety issues.}
\end{figure}

\section{Methodology}

This literature review aims to comprehensively examine the trustworthiness of Multilingual Language Models (MLLMs) in collaborative systems. We followed a systematic review process inspired by methodologies from prior works \cite{pater2021standardizing, aghajari2023reviewing}, adapting them to explore the socio-technical challenges and opportunities presented by MLLMs in HCI.

To identify relevant studies, we performed our search in the ACM Digital Library and Google Scholar. These databases were selected for their extensive coverage of research in HCI, CSCW, CMC and AI, which are central to the development and deployment of MLLMs. The scope of our search includes papers published within the last decade to ensure contemporary relevance.

\begin{table}[ht]
  \caption{Categorization of Citations by Key Attributes and Discussion Themes}
  \label{tab:citations}
  \begin{tabular}{ |>{\raggedright\arraybackslash}p{4.5cm}|>{\raggedright\arraybackslash}p{3.5cm}|>{\raggedright\arraybackslash}p{4.5cm}| }
    \toprule
    Attribute/Theme & Citation & Comments \\
    \midrule
    Data Quality & \cite{10275753}, \cite{10448361}, \cite{10411566}, \cite{ahuja2023mega},\cite{chen2023monolingual}, \cite{zhang2023don}, \cite{choudhury2023ask}, \cite{ali2023tokenizer}, \cite{bommasani2021opportunities}, \cite{kassner2021multilingual}, \cite{muller2020being}, \cite{pfeiffer2020unks}, \cite{barbieri2021xlm} & Addresses issues of data bias and representation across languages \\
     \hline
    Dependability \& Operability & \cite{gangavarapu2023llms}, \cite{10293675}, \cite{10193533}, \cite{shen2024language}, \cite{yong2023low}, \cite{gangavarapu2023llms}, \cite{choudhury2021linguistically}, \cite{winata2021language}  & Focuses on model reliability and consistency in multilingual applications\\
     \hline
    Human-Centered Design & \cite{10193533}, \cite{10293675}, \cite{tokayev2023ethical}, \cite{choudhury2021linguistically}, \cite{cahyawijaya2024llms}, \cite{adebara2022serengeti} & Emphasizes cultural sensitivity and inclusivity in model design \\
     \hline
    Human Oversight & \cite{10293675}, \cite{floridi2018soft}, \cite{weidinger2022taxonomy}, \cite{tokayev2023ethical}, \cite{jobin2019global}, \cite{prunkl2021institutionalizing},\cite{kwak2024bridging}, \cite{raji2020closing}, \cite{gangavarapu2023llms}, \cite{whittlestone2019role}, \cite{amershi2019guidelines}, \cite{liang2021towards} & Discusses ethical considerations and the role of human oversight in model development \\
     \hline
    Real-World Applications & \cite{tanwar2023multilingual}, \cite{tanwar2023multilingual}, \cite{castillo2021dark}, \cite{roslan2023rise}, \cite{10291329}, \cite{kusal2023systematic}, \cite{yenduri2024gpt}, \cite{gangavarapu2023llms}, \cite{rasiah2024one}, \cite{HULL2016158}, \cite{chang2024survey}, \cite{dankwa2021proposed}, \cite{pressman2024ai}, \cite{pan2024feedback}, \cite{NEURIPS2022_ce9e92e3} & Considers practical applications in areas like healthcare and legal services \\
     \hline
    Socio-Political Impact & \cite{mendoncca2023simple}, \cite{C2021Artificial}, \cite{Cooper2023Bridging}, \cite{Chin2023Linguistic},\cite{Petr0Large}, \cite{bozdag2023aismosis}, \cite{T2021Mind}, \cite{unCountriesAdopt}, \cite{worldbankTechnologyAdoption}, \cite{Inuwa2023FATE}, \cite{Caianiello2021Dangerous}, \cite{Naseeb2020AI}, \cite{C2017framework} & Examines MLLM’s potential impact on education, healthcare, and employment \\
  \bottomrule
\end{tabular}
\end{table}

\subsection{Search Strategy}

Our search used a set of keywords crafted to capture the intersection of MLLM and collaborative work. Keywords included 'Multilingual Language Models', 'MLLMs in Collaboration', 'AI for multilingual collaboration', and 'Trustworthiness in AI', among others. This set of keywords was chosen to encompass both the technical and human-centered aspects of MLLM. Additional terms, such as Human-Centered AI, Data Quality in MLLM, and Ethical AI deployment, were iteratively refined to broaden the scope without compromising specificity. The rationale for using these keywords lies in the multifaceted nature of trustworthiness in MLLMs. For instance, "Data Quality" captures the need for accurate and unbiased data to support equitable AI outcomes, while "Human-Centered AI" highlights the importance of designing models that cater to diverse user groups. These keywords ensure that we not only cover technical robustness, but also address the social and cultural implications of MLLMs, which are critical for their integration into collaborative systems \cite{tokayev2023ethical, bender2021dangers}.

In the ACM Digital Library, we restricted our selection by filtering for papers that included these keywords in the abstract and limited the publication date to the past five years. This restriction was applied due to the relatively recent surge in interest and development in large language models (LLMs), which have gained significant attention in the past few years, marking a shift in their widespread application and research focus.

The articles were also selected based on the following inclusion criteria: First, they needed to focus on MLLMs or related technologies within contexts of multilingual, multicultural collaboration, and communication. Furthermore, studies were required to address one or more of the four critical attributes: data quality, dependability and availability, human-centered design, and human oversight, especially in how these attributes relate to fostering effective cross-linguistic and cross-cultural interaction. Lastly, we include studies exploring how MLLMs contribute to or impact collaborative systems, with an emphasis on improving or mediating communication across diverse cultural and linguistic backgrounds.

The ACM Digital Library search yielded 103 results, which were further narrowed based on relevance to our focus on multilingual collaboration. Given that LLM is a relatively new research domain, we also considered several arXiv preprints that aligned closely with our ideas. These preprints provide valuable information on emerging trends and challenges in the field that are not yet fully captured in the peer-reviewed literature.

\subsection{Data Extraction and Categorization}

For each selected article, we extracted data on the publication venue, year, research methodology, and findings related to our four focus areas. These attributes—\textbf{Data Quality}, \textbf{Dependability and Operability}, \textbf{Human-Centered Design}, and \textbf{Human Oversight}—were chosen based on their prevalence and relevance in assessing trustworthiness in AI systems. 

Each of these attributes serves a specific purpose in understanding the strengths and limitations of MLLMs within collaborative systems. Data quality is crucial because it directly influences the accuracy and fairness of the model, particularly for low-resource languages, which are often underrepresented in pre-training datasets. Ensuring high data quality helps mitigate bias and improves inclusivity, which is essential for equitable collaboration. Dependability and Operability are crucial, as they focus on the model's reliability and practical application in real-world settings. For MLLMs to function effectively within the CSCW and CMC, they must be reliable and adaptable to diverse operational environments, facilitating sustainable and consistent communication.

Human-Centered Design emphasizes the need for MLLMs to be accessible and intuitive across a variety of linguistic and cultural backgrounds. By prioritizing this attribute, the review acknowledges that effective collaboration requires AI tools that are user-friendly and responsive to the needs of all users, regardless of their technical expertise. Lastly, Human Oversight is essential to ensure ethical deployment and maintaining user trust. This attribute reflects the growing awareness of the ethical implications of AI, underscoring the importance of transparency, accountability, and bias mitigation in MLLMs.

The four attributes—Data Quality, Dependability, Human-Centered Design, and Human Oversight—form a comprehensive framework for evaluating the impact of MLLMs in socially mediated environments. This framework addresses not only the technical robustness of MLLMs but also the broader social and cultural implications that are essential for the HCI community. To conduct this literature review, we employed a systematic methodology that began by constructing a review corpus, drawing from relevant literature in the ACM Digital Library and arXiv to encompass studies focused on multilingual and multicultural collaboration. As depicted in Figure 2, this process included defining the key attributes that provide a robust foundation for assessing MLLMs’ effectiveness in fostering cross-linguistic and cross-cultural interactions.

The final stage involved a detailed analysis of how these attributes impact collaborative interactions in real-world contexts, shedding light on how MLLMs facilitate or hinder communication across diverse linguistic and cultural landscapes. By evaluating MLLMs through this multidimensional framework, our review seeks to offer a holistic perspective on their strengths and limitations, particularly in supporting inclusive and effective collaboration across languages and cultures. This structured approach not only addresses technical aspects but also emphasizes the social dimensions, ensuring a balanced understanding of MLLMs’ role in global collaboration.

\begin{figure}[ht]
  \centering
  \includegraphics[width=\linewidth]{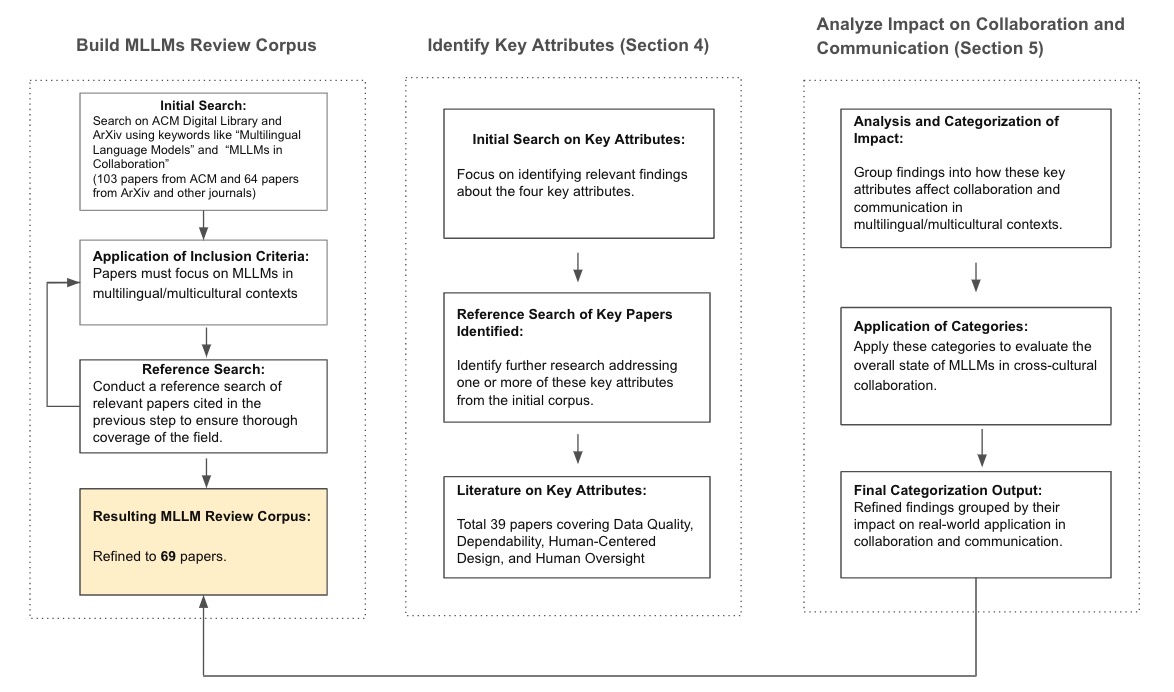}
  \caption{Systematic methodology for MLLM literature review process.}
  \Description{A flowchart detailing the literature review methodology, broken down into three main stages: Building the MLLM review corpus, identifying key attributes, and analyzing the impact on collaboration and communication.}
\end{figure}

\section{Results}

\subsection{Data Quality}

Data quality plays a fundamental role in shaping the robustness and effectiveness of MLLMs, especially in multilingual and multicultural contexts, where languages have varying levels of resource availability. Comprehensive, accurate, and culturally diverse datasets are essential to ensure that these models work equitably across both high- and low-resource languages. The shift from traditional machine translation to a more holistic, multilingual engagement supports nuanced communication and collaboration across cultures \cite{nie2024decomposed, qin2024multilingual}. However, with approximately 7,000 spoken languages and 300 writing systems worldwide, only a fraction is well represented in the training datasets for MLLMs \cite{qin2024multilingual}. This limited representation favors high-resource languages like English, Mandarin, and Spanish, while marginalizing low-resource languages \cite{chen2023monolingual, zhang2023don, choudhury2023ask}.

Bimagambetova et al. \cite{10275753} highlight the impact of inadequate data for languages like Kazakh, revealing that insufficient training data leads to model underperformance, which not only affects accuracy but also introduces biases that undermine cross-cultural communication and collaboration. Kassner et al. \cite{kassner2021multilingual} further demonstrate that multilingual models like mBERT exhibit language biases, achieving higher performance on well-resourced languages while performing poorly on others, underscoring the need for balanced data representation.

Pfeiffer et al. \cite{pfeiffer2020unks} tackle the challenge of adapting MLLMs to new and underrepresented scripts. They propose data-efficient methods, including matrix factorization, that capitalize on existing latent knowledge within the model’s embedding matrix. Their approach also leverages lexically overlapping tokens to improve performance on low-resource languages and those written in scripts previously unseen by the model. This method is particularly valuable for enhancing model adaptability and accuracy in truly multilingual applications, enabling MLLMs to bridge gaps in language resources and providing substantial performance improvements for low-resource languages with unique scripts.

To address these data challenges, researchers have also focused on synthetic data generation and model fusion techniques to enhance low-resource language support. Huang et al. \cite{10448361} propose a model fusion approach to maximize data utility, bolstering robustness by compensating for data scarcity. Bhavya et al. \cite{10411566} demonstrate that synthetic data can improve resilience and robustness in niche domains where natural language data is limited. Barbieri et al. \cite{barbieri2021xlm} present a domain-specific approach, XLM-T, which pre-trains on multilingual social media data (Twitter), showing how tailored datasets can enhance MLLM performance on sentiment analysis across multiple languages.

Ahuja et al. \cite{ahuja2023mega} stress the importance of comprehensive multilingual evaluation to ensure consistent performance across 70 languages, indicating that robust data quality is crucial for trustworthy MLLM outputs in diverse multicultural settings \cite{bender2021dangers, adebara2022serengeti}.

By addressing these data quality challenges, MLLMs can better support equitable multilingual collaboration and enhance the inclusivity of AI-driven communication tools across diverse linguistic and cultural backgrounds.

\subsection{Dependability and Operability}

Dependability and operability are essential attributes for MLLMs, particularly in high-stakes applications requiring consistent, reliable performance across multiple languages, such as healthcare and cross-cultural communication. Ensuring that these models operate effectively across varied and dynamic contexts supports the resilience required for effective multilingual collaboration. Gangavarapu \cite{gangavarapu2023llms} and Gangavarapu et al. \cite{gangavarapu2023llms} explore the use of MLLMs in healthcare, demonstrating how dependable and robust models can enhance service delivery in low-resource settings where reliability is crucial for successful outcomes. In these scenarios, robustness is defined by the model’s capacity to handle diverse data inputs and maintain consistent performance, even in high-stakes environments that require adaptability across languages. 

Winata et al. \cite{winata2021language} further contribute by illustrating how general-purpose models like GPT and T5 can achieve cross-lingual tasks with minimal data through few-shot learning. Their findings reveal that, with a few examples in English, these models can accurately generalize predictions to non-English samples. This few-shot cross-lingual capability is particularly valuable in low-resource settings, where such models exhibit reliable performance across languages without extensive retraining, thereby enhancing the operability and adaptability of MLLMs in multilingual applications.

A dependable MLLM must also maintain fairness and equity across both high- and low-resource languages, as highlighted by Choudhury and Deshpande \cite{choudhury2021linguistically}. Their research emphasizes that consistency across languages is a critical aspect of fairness, proposing Rawlsian fairness principles to prioritize underrepresented languages in model selection. By considering these fairness principles, models can achieve more balanced and equitable performance, reinforcing dependability across diverse linguistic contexts.

Dependability is further strengthened through multimodal data integration, as discussed by Mani and Namomsa \cite{10293675}. Incorporating visual, auditory, and textual inputs enhances MLLMs’ adaptability to real-world situations, broadening their utility in diverse settings. This multimodal approach is particularly relevant in multilingual applications, where, as Lee et al. \cite{10193533} demonstrate, vocabulary sharing improves translation robustness and consistency, yielding more accurate outputs in cross-cultural interactions. Additionally, research on multilingual security vulnerabilities by Shen et al. \cite{shen2024language} and Yong et al. \cite{yong2023low} underscores the necessity for resilient, dependable models that mitigate language-specific risks, further supporting reliable and equitable communication across linguistic boundaries.

\subsection{Human-Centered Design}

Human-centered design in MLLMs is essential for ensuring accessibility, usability, and responsiveness to the diverse needs of multilingual and multicultural user bases. Prioritizing non-technical end-users in the design process enables MLLMs to foster effective communication across varied linguistic backgrounds. A compelling example is provided by Lee et al. \cite{10193533}, whose work on sign language translation illustrates the direct benefits of MLLMs for the deaf community, highlighting the transformative potential of these models in improving quality of life for diverse users. This underscores the critical importance of inclusivity in MLLM development, where addressing linguistic accessibility has a tangible impact on everyday communication.

Adebara et al. \cite{adebara2022serengeti} further demonstrate the significance of human-centered design by focusing on low-resource language challenges. Their work on the SERENGETI framework emphasizes the need for MLLMs that are sensitive to linguistic diversity, particularly for underrepresented languages. By developing models that are both inclusive and adaptable, MLLMs can serve a broader audience and maintain dependability across language communities with varying resource levels, thus promoting equity in multilingual digital spaces.

Beyond individual user needs, human-centered design in MLLMs addresses broader societal and practical applications, as emphasized by Mani and Namomsa \cite{10293675}. By incorporating human-centered principles, MLLMs can support a broad range of societal needs beyond technical requirements. This inclusive approach enables MLLMs to be accessible and usable by people from various cultural and linguistic backgrounds, which is particularly crucial in fields such as education and healthcare. For instance, Gangavarapu \cite{gangavarapu2023llms} highlights the potential of user-friendly MLLMs in enhancing service delivery in low-resource settings, demonstrating how human-centered MLLMs can play a vital role in bridging gaps in essential services.

The social and political implications of human-centered MLLM design are substantial. Tokayev et al. \cite{tokayev2023ethical} and Choudhury and Deshpande \cite{choudhury2021linguistically} argue that failing to accommodate diverse linguistic needs risks reinforcing educational and economic disparities by prioritizing high-resource languages. Cahyawijaya et al. \cite{cahyawijaya2024llms} illustrate how underrepresented languages are often deprived of the educational benefits provided by well-resourced MLLMs. Addressing these gaps with a human-centered design approach allows MLLMs to offer equitable access to digital resources, fostering enhanced communication and collaboration across diverse linguistic landscapes.

\subsection{Human Oversight}

Human oversight plays a significant role in guiding the ethical development and deployment of MLLMs, ensuring these models remain aligned with human values and societal welfare across multilingual and multicultural contexts. This oversight encompasses ethical AI development, consistent monitoring, regulatory compliance, and human-in-the-loop approaches, which are essential for managing the potential impacts of MLLMs. As Mani and Namomsa \cite{10293675} highlight, ethical considerations are fundamental in AI development to prevent the reinforcement of biases and unjust outcomes. Addressing these disparities aligns with the imperatives outlined by Tokayev et al. \cite{tokayev2023ethical}, who emphasize the importance of developing inclusive AI systems that respect and uplift marginalized languages and cultures. Ensuring the dependability and inclusivity of MLLMs requires that these ethical foundations be a priority, supporting equitable interactions across diverse linguistic communities.

Floridi et al. \cite{floridi2018soft} advocate for an ethical framework grounded in principles like beneficence, nonmaleficence, and justice—principles crucial for designing AI systems that genuinely serve societal needs. Jobin et al. \cite{jobin2019global} add to this by reviewing global AI ethics guidelines, offering a foundation for embedding human values within AI. This perspective aligns with Prunkl et al. \cite{prunkl2021institutionalizing}, who stress the importance of embedding ethics into AI through impact assessments, thereby ensuring that MLLMs are responsibly deployed across diverse cultural contexts.

The potential social and political effects of MLLMs necessitate robust oversight mechanisms to prevent unintended consequences. Kwak et al. \cite{kwak2024bridging} caution that MLLMs without diverse linguistic data risk reinforcing social inequalities and diminishing cultural diversity. Raji et al. \cite{raji2020closing} advocate for AI auditing and regulatory frameworks, especially in high-stakes sectors like healthcare, as demonstrated in Gangavarapu’s \cite{gangavarapu2023llms} work. Whittlestone et al. \cite{whittlestone2019role} propose that AI systems be assessed based on their risk level and societal impact—an approach that is particularly relevant in applications affecting socio-political landscapes.

Incorporating human-in-the-loop systems further strengthens MLLM oversight, enhancing reliability and trust by embedding continuous user feedback. Amershi et al. \cite{amershi2019guidelines} provide critical guidelines for human-AI interaction, which are essential for aligning MLLMs with user expectations and ethical standards. Additionally, Liang et al. \cite{liang2021towards} propose methodologies to detect and address social biases within language models, a necessary step for fair and ethical MLLM deployment.

Integrating human oversight throughout the MLLM lifecycle enables developers to ensure that these models not only comply with ethical standards and legal frameworks but also positively contribute to multilingual and multicultural collaboration. Effective oversight mitigates socio-political risks, builds trust, and ensures that MLLMs serve as reliable tools in diverse, computer-mediated communication environments.

\section{Discussion}

MLLMs offer substantial advances over traditional language technologies, yet require a fundamentally different approach that extends beyond mere translation capabilities. These models are not just tools for converting text from one language to another; they must navigate the intricate cultural, contextual, and linguistic landscapes that shape human communication. As Bender et al. (2021) highlight, language models are not inherently neutral, and risk-amplifying biases embedded in the data used for their training. This tendency can result in cultural misunderstandings, stereotypes, and a lack of representation of low-resource languages if not managed with careful consideration of social and cultural contexts \cite{bender2021dangers}.

The need for MLLMs that are capable of supporting meaningful cross-cultural interactions is especially critical in socially mediated environments. In fields like Computer-Supported Cooperative Work (CSCW), interactions often involve participants from diverse linguistic and cultural backgrounds. For MLLMs to be truly effective in these settings, they must go beyond accurate language conversion to address the broader dimensions of cultural sensitivity, inclusivity, and contextual understanding. 

A significant gap in current MLLM development is insufficient attention to multicultural, as opposed to solely multilingual, collaboration. Although many models are adept at addressing language barriers, they often overlook how cultural contexts, values, and social norms influence communication. This limitation can hinder the effectiveness of MLLMs in diverse environments, as models that fail to account for cultural differences may inadvertently reinforce stereotypes or misunderstandings. For example, recent efforts like SERENGETI, which incorporates cultural context for African languages, underscore the importance of designing MLLMs that reflect and respect the cultural diversity of the users they serve.

Furthermore, the potential for MLLMs to function as collaborative agents – rather than mere translation tools – opens up new possibilities for their use in multicultural team-based settings. Unlike machine translation systems that focus on text conversion, MLLMs can provide contextually appropriate responses, assist decision-making, and facilitate problem solving across linguistic and cultural boundaries. This expanded role aligns with the needs of team science, where collaboration frequently spans diverse linguistic and cultural contexts.

The socio-political and ethical implications of deploying MLLMs also warrant careful consideration. These models can influence a wide range of social domains, including education, economics, and legal systems. In education, for example, MLLMs could bridge or exacerbate disparities. Although they offer the potential to improve language learning and educational content, their tendency to favor high-resource languages can disadvantage speakers of underrepresented languages \cite{cahyawijaya2024llms}. Similarly, in economic terms, the automation capabilities of MLLMs can impact job markets in ways that are beneficial and disruptive, underscoring the need for strategies that mitigate potential job displacement while fostering new opportunities in AI-driven industries.

Ethical oversight plays a critical role in addressing these challenges. As suggested by Raji et al. \cite{raji2020closing}, regulatory frameworks and continuous monitoring are essential to mitigate the risks of MLLMs exacerbating inequalities or reinforcing biases. This includes implementing human-in-the-loop systems that ensure MLLMs remain adaptable and aligned with human values. By embedding ethical considerations throughout the MLLM lifecycle, developers can safeguard against potential harms and promote the responsible deployment of these models in multicultural and multilingual contexts.

\subsection{Real-World Impact on End-Users}

MLLMs are integral in shaping the future of human-computer interaction, especially in multilingual contexts \cite{tanwar2023multilingual, tanwar2023multilingual}. However, the prevalent training biases toward high-resource languages often result in systems that fail to serve a global user base effectively. This bias has particularly severe consequences for speakers of minority languages, which represent the majority of human languages but are underrepresented in the datasets used to train MLLMs. If these languages disappear from digital platforms and AI models, it will lead to the erosion of cultural heritage and the loss of knowledge encoded in these languages. Unlike English and Spanish, which share similar roots, many minority languages have completely distinct linguistic structures. This lack of linguistic and cultural representation introduces challenges in training models to understand these languages and undermines the development of inclusive AI systems.

The impact of neglecting minority languages extends beyond mere technological inconvenience, posing a threat to linguistic diversity and cultural survival. In scenarios where these languages disappear, AI systems would be further skewed toward dominant languages, resulting in systems that lack the ability to understand, communicate, or translate for significant portions of the global population. This would exacerbate linguistic inequalities, making technology less accessible for many communities. As a result, users who speak minority languages will face limited usability and increased frustration when interacting with AI-driven applications such as virtual assistants or chatbots of customer service, which will struggle to process their input \cite{castillo2021dark, roslan2023rise,10291329}. Such limitations diminish trust in AI technologies and hinder their widespread adoption, especially in regions where diverse languages are spoken \cite{kusal2023systematic, yenduri2024gpt}.

The application of MLLMs in critical services such as healthcare, legal and government services has enormous potential to improve efficiency and accessibility \cite{gangavarapu2023llms, rasiah2024one}. However, when these systems fail to address the needs of all language speakers equally, they can exacerbate existing inequalities. For example, in healthcare, miscommunication due to language processing errors could lead to improper diagnoses or treatment plans, disproportionately affecting non-native speakers \cite{HULL2016158}. This underscores the urgent need for MLLMs that are equipped to handle minority languages as effectively as high-resource languages.

In legal settings, the stakes are similar. MLLMs that help to understand and translate legal documents must do so with high accuracy to prevent legal misinterpretations that could unfairly impact non-English speakers \cite{chang2024survey}. Developing models that handle minority languages with the same precision and nuance as major languages is critical to ensuring justice and equity in legal processes \cite{dankwa2021proposed}. To address these challenges, there is a growing emphasis on developing human-centered AI systems that prioritize the needs and nuances of human interactions. This involves designing MLLMs that are not only linguistically inclusive but also aware of cultural and contextual subtleties \cite{pressman2024ai}.

By incorporating principles of HCI and user experience design, developers can create more empathetic and adaptive systems. For instance, feedback loops where users can report and correct misunderstandings can improve the accuracy of MLLMs over time \cite{pan2024feedback}. Moving forward, interdisciplinary collaborations between linguists, cultural scholars, and AI researchers will be essential to ensure that these models respect and enhance linguistic diversity. This approach will not only improve the functionality of AI systems, but will also ensure that they are deployed in a way that values cultural and linguistic diversity at a deep level \cite{NEURIPS2022_ce9e92e3}. By emphasizing human-centered design and inclusive development practices, the future of HCI can evolve to meet the needs of a truly global user base, making technology accessible and beneficial for all, regardless of language or cultural background.

\subsection{Policy and Legal Considerations}

The advancement of MLLMs requires a concerted effort to establish robust policy and legal frameworks that ensure the equitable dissemination and inclusive development of these technologies \cite{mendoncca2023simple}. Current frameworks often lack provisions to ensure that the development of AI technologies includes all linguistic groups. To remedy this, policies must mandate the inclusion of diverse linguistic datasets in AI training processes and enforce transparency in the construction and deployment of these models \cite{C2021Artificial}. Existing processes for ensuring the safety of large language models are inadequate and fail to provide democratic legitimacy \cite{Cooper2023Bridging}. Artificial languages offer promising avenues for more equitable AI deployment and the potential to bridge linguistic divides \cite{Chin2023Linguistic}. Furthermore, large language models could enhance the epistemic outcomes of democracy and contribute to sustainability by improving the accessibility and availability of expert knowledge \cite{Petr0Large}. Legal standards should also be introduced to protect against the misuse of AI in ways that could exacerbate social inequalities.

The need for robust policy and legal frameworks to address the challenges posed by MLLMs is evident as their influence expands globally\cite{bozdag2023aismosis}. These models must be developed with an eye toward inclusivity, ensuring that linguistic and cultural diversity is respected and enhanced rather than undermined. Addressing the disparities caused by MLLMs requires significant policy and legal considerations, particularly when considering the roles of global organizations and governments in developed and developing countries. International entities such as the United Nations (UN) and the World Bank have recognized the profound impact of technology, including AI, on societal and economic development. These organizations emphasize the importance of equitable technology deployment to ensure that advancements in AI contribute positively across all sectors of society \cite{T2021Mind}.

The UN has taken steps to create a global agreement on the ethics of artificial intelligence adopted by UNESCO member states. This historic agreement aims to ensure that AI development benefits humanity while addressing potential threats such as privacy violations, bias, and increased surveillance. It emphasizes the necessity of constructing legal frameworks to guide the ethical development of AI technologies \cite{unCountriesAdopt}. Similarly, the World Bank highlights the importance of technology adoption by firms in developing countries as crucial to economic growth and resilience. It stresses that technological disparities between firms in developed and developing countries can exacerbate income inequality and slow economic progress. Policies that facilitate the adoption of advanced technologies in developing countries are seen as imperative to reducing these disparities and promoting equitable economic development \cite{worldbankTechnologyAdoption}.

Both organizations advocate for policies that ensure inclusive technology development that respects human rights and promotes diverse linguistic and cultural participation. This includes encouraging the use of open source technologies to allow communities to develop solutions that address their specific needs, potentially reducing dependence on major tech conglomerates and promoting technological self-reliance \cite{unCountriesAdopt, worldbankTechnologyAdoption}.

Publications emphasize the need for policies reflecting fairness, accountability, transparency, and ethics (FATE) in AI, advocating for the involvement of diverse stakeholders to ensure that AI technology is accessible and equitable \cite{Inuwa2023FATE}. These discussions are critical as they highlight the neglect of local knowledge and cultural pluralism, especially in the Global South. The discourse examines the inclusivity of global AI ethics and the need for democratic legitimacy in technology deployment \cite{C2021Artificial, Cooper2023Bridging}.

To effectively address the challenges posed by MLLMs and leverage their potential for positive impact, it is crucial that policy and legal frameworks are established that promote the inclusion of diverse linguistic datasets, ensure transparency in AI deployments, and protect against AI misuse. Such measures are crucial in preventing the exacerbation of social inequalities \cite{Cooper2023Bridging, Caianiello2021Dangerous}. These frameworks should also support the development of technologies that are accessible to and inclusive of all linguistic communities, particularly those in developing regions, fostering environments that encourage the development of AI solutions tailored to the needs of diverse linguistic and cultural groups \cite{Naseeb2020AI, C2017framework}. The integration of AI in democratic processes and its potential to improve deliberative democracy also underlines the need for AI technologies that are inclusive and capable of supporting sustainable development \cite{Naseeb2020AI}.

\subsection{Future Directions}

To advance the development and deployment of MLLMs, the CSCW and HCI communities should explore cross-disciplinary collaborations that involve AI researchers, linguists, social scientists, and cultural theorists. This interdisciplinary approach is essential for creating MLLMs that are not only technically advanced but also culturally informed and socially responsible. These collaborations can help bridge the gap between technical innovation and real-world applicability, ensuring that MLLMs are designed with an understanding of cultural diversity and the complexities of communication in multilingual and multicultural settings.

Based on the insights of this review of the literature, future research should prioritize the development of tools and metrics that go beyond traditional evaluation methods, such as BLEU scores. New frameworks should be designed to assess how well MLLMs facilitate collaboration, support inclusive communication, and respect cultural nuances. For example, MLLMs could be evaluated on their ability to mediate interactions in cross-cultural teams, ensuring that linguistic and cultural differences are navigated effectively in real-world settings. Such metrics could provide valuable insights into the models' effectiveness in fostering global collaboration and reducing miscommunication, making MLLMs an essential resource for enhancing cross-cultural teamwork.

Moreover, this literature review highlights the need for participatory design practices and user feedback loops to ensure that MLLMs are developed in a way that aligns with the needs and expectations of diverse communities \cite{cao2024towards}. Engaging end-users in the design process can lead to models that are more adaptable, accessible, and relevant to a global user base. Participatory methods are particularly important for identifying the unique challenges faced by low-resource language speakers and marginalized cultural groups, whose needs are often overlooked in AI development. Workshops, task forces, and collaborative projects should prioritize discussions on ethical deployment, real-world application challenges, and best practices for integrating MLLMs into various collaborative systems.

Even though multilingual communities that use machine translation to overcome language barriers are increasing, we still lack a complete understanding of how machine translation affects communication. One avenue for future research could involve redesigning and refining existing CSCW machine translation experiments, tailoring them to the specific challenges and opportunities presented by MLLMs. For example, Yamashita et al. \cite{yamashita2006effects} examined the impact of machine translation on communication by engaging participants from China, Korea, and Japan in referential tasks. Their findings highlighted disruptions in lexical entrainment and inconsistencies in referring expressions caused by asymmetries in machine translations. These challenges made communication less efficient, as translations often failed to consistently echo terms, disrupting natural conversational flow.

Similarly, Hautasaari et al. \cite{hautasaari2010machine} conducted a laboratory experiment involving intercultural distributed groups using machine translation-mediated chat in a trading game scenario. Participants from Finland and Japan used machine translation as their main communication tool. Their study found that, despite language barriers, machine translation had a positive effect on social and relational communication, increasing the number of positive socioemotional messages and improving overall group performance. These findings suggest that machine translation can facilitate more effective teamwork in intercultural groups, although challenges remain in the consistency and quality of translations.

Building upon this research, the application of MLLMs in similar scenarios offers a promising direction. MLLMs can address some of the limitations identified by Yamashita et al. and Hautasaari et al. by providing more context-aware and culturally sensitive translations that could reduce inconsistencies and asymmetries in communication. Future work could involve testing how MLLMs perform in real-world collaborative environments where linguistic and cultural diversity is pivotal. This approach can help evaluate the ability of MLLMs to facilitate more effective teamwork between languages and cultures, refining existing methodologies to optimize cross-linguistic collaboration \cite{yamashita2006effects, hautasaari2010machine}. Drawing on past CSCW experiments, researchers can develop more robust frameworks to assess the impact of MLLMs on cross-cultural communication, thus bridging the gap between human-centric AI and machine translation.

In addition, the CSCW and HCI communities should work to strengthen connections with the AI and machine learning communities, creating opportunities for joint conferences, workshops, and collaborative research projects. Conferences that bring together AI and CSCW researchers can facilitate the exchange of ideas and foster new collaborations focused on the intersection of AI, multilingualism, and cross-cultural communication. Joint sessions at major AI and machine learning conferences can also provide a platform to discuss the ethical implications of MLLMs and develop best practices for their deployment in diverse cultural contexts.

Future work should also focus on theory-building and methodology design in the context of MLLMs. Although much research has been conducted on personalization and user experience, there is still a lack of robust methodologies to evaluate the impact of MLLM on multicultural collaboration. The development of new theories that integrate cultural and linguistic diversity into AI design is crucial for guiding future research and ensuring that MLLMs are capable of fostering effective communication across a broad spectrum of cultural contexts. As a community, CSCW researchers should advocate for the development of such theories, working together to build and design methodologies that ensure MLLMs contribute positively to global communication.

In conclusion, the research community must take a proactive role in shaping the future of MLLMs by addressing the unique challenges of multicultural and multilingual collaboration. By fostering interdisciplinary collaborations, developing new evaluation metrics, and engaging in participatory design, the CSCW and HCI communities can contribute to the development of MLLMs that bridge linguistic and cultural divides. Through joint conferences, new workshops, and the creation of dedicated research initiatives, we can ensure that MLLMs are not only technically capable but also socially and culturally responsible, ultimately promoting more inclusive and effective collaboration on a global scale.

\section{Limitations}

This survey has a few limitations that should be acknowledged. Firstly, while our aim has been to provide a comprehensive review of MLLMs, the rapidly evolving nature of this field means that our selection of articles included both peer-reviewed publications and arXiv preprints. The inclusion of preprints is necessary to capture the latest advancements and emerging discussions, as MLLMs represent a relatively new area of research with ongoing developments. However, these preprints have not undergone formal peer review, which can affect the reliability of some insights compared to fully peer-reviewed studies.

Furthermore, this survey has been conducted with a focus on the literature available in English, which may bias the findings toward English-centric perspectives on multilingual and multicultural challenges. Although English is widely used in academic publications, this linguistic limitation may inadvertently exclude research conducted in other languages, particularly those that focus on low-resource languages from non-English-speaking regions.

\section{Conclusion}

In this survey, we examined the trustworthiness and potential of MLLMs in supporting multilingual and multicultural collaboration. By focusing on four critical attributes—\textbf{Data Quality}, \textbf{Dependability and Operability}, \textbf{Human-Centered Design}, and \textbf{Human Oversight}—we have established a framework that highlights both the capabilities and limitations of MLLMs in fostering cross-cultural communication. This structured approach underscores the unique challenges faced by MLLMs, including disparities in data representation for low-resource languages, the need for culturally aware model design, and the essential role of ethical and regulatory oversight in ensuring responsible deployment.

Our review reveals that while MLLMs hold considerable promise for enabling more inclusive and effective communication in computer-mediated environments, significant efforts are still required to align these models with the needs of a truly global user base. The sociopolitical and ethical implications of MLLMs, particularly in critical areas such as education, healthcare, and economic development, underscore the necessity of a human-centered approach. MLLMs must prioritize cultural sensitivity and ethical deployment to mitigate the risks of reinforcing biases and exacerbating inequalities among diverse linguistic and cultural groups.

As MLLMs continue to evolve, it is essential for the research community, including the broader HCI and AI communities, to prioritize interdisciplinary collaborations that bridge technical development with cultural and social expertise. By fostering these partnerships, MLLMs can be developed to be not only linguistically inclusive but also capable of navigating complex cultural contexts. Such collaborative efforts are key to enhancing the trustworthiness, accessibility, and relevance of AI technologies in global settings.

Future research should focus on designing and evaluating MLLMs as tools that enhance cross-cultural collaboration and effective communication in diverse multilingual contexts. This includes creating frameworks that incorporate cultural sensitivities, promoting inclusivity, and developing evaluation metrics that go beyond the accuracy of translation. New metrics should assess how well MLLMs facilitate understanding, foster trust, and enable equitable participation in multilingual teams, positioning these models as collaborative agents rather than mere translation tools. Addressing these challenges will be essential to establish MLLM as integral resources that support global communication and foster meaningful interactions across cultural divides.

In conclusion, while this survey offers a foundational understanding of the current landscape and critical issues in MLLMs, ongoing research, policy development, and cross-disciplinary collaboration will be vital to fully harnessing their potential in a manner that is ethically responsible and culturally sensitive. By addressing these complexities, MLLMs can become powerful tools to bridge cultural and linguistic gaps, promoting more inclusive, effective, and trustworthy AI-driven communication throughout the world.

\bibliographystyle{ACM-Reference-Format}
\bibliography{cscw.bib}

\end{document}